\pdfoutput=1
\documentclass[11pt]{article}

\usepackage[]{acl}

\usepackage{times}
\usepackage{latexsym}

\usepackage[T1]{fontenc}
\usepackage[utf8]{inputenc}

\usepackage{microtype}

\usepackage{inconsolata}

\usepackage{graphicx}
\usepackage{booktabs}
\usepackage{multirow}
\usepackage{xcolor} 
\usepackage{listings}
\usepackage{hyperref}
\usepackage{multicol} % for multiple columns
\usepackage{lipsum} % for dummy text
\usepackage{amsmath}
\usepackage{amssymb}
\usepackage{algorithm}
\usepackage{algorithmicx}
\usepackage{algpseudocode}
\usepackage{array}
\usepackage{booktabs}
\usepackage{arydshln}

\title{Make Every Penny Count: Difficulty-Adaptive Self-Consistency for Cost-Efficient Reasoning}

\begin{document}
\author {
    % Authors
    \textbf{Xinglin Wang}\textsuperscript{\rm 1}, \hspace{0cm}
    \textbf{Shaoxiong Feng}\textsuperscript{\rm 2}, \hspace{0cm}
    \textbf{Yiwei Li}\textsuperscript{\rm 1}, \hspace{0cm} 
    \textbf{Peiwen Yuan}\textsuperscript{\rm 1}, \hspace{0cm} 
    \textbf{Yueqi Zhang}\textsuperscript{\rm 1}, \hspace{0cm} \\
    \textbf{Chuyi Tan}\textsuperscript{\rm 1}, \hspace{0cm}
    \textbf{Boyuan Pan}\textsuperscript{\rm 2}\textbf{,} \hspace{0cm} 
    \textbf{Yao Hu}\textsuperscript{\rm 2}\textbf{,} \hspace{0cm} 
    \textbf{Kan Li}\textsuperscript{\rm 1}\footnotemark[2] \\
    \textsuperscript{\rm 1} School of Computer Science, Beijing Institute of Technology \\
    \textsuperscript{\rm 2} Xiaohongshu Inc \\
    \texttt{\{wangxinglin,liyiwei,peiwenyuan, zhangyq, tanchuyi, likan\}@bit.edu.cn} \\
    \texttt{\{shaoxiongfeng2023\}@gmail.com} \  \texttt{\{panboyuan,xiahou\}@xiaohongshu.com}
    %\texttt{\{shaoxiongfeng2023,whd.thu\}@gmail.com}  \quad \texttt{\{panboyuan\}@xiaohongshu.com}
}

\maketitle

\renewcommand{\thefootnote}{\fnsymbol{footnote}} 
\footnotetext[2]{Corresponding author.} 

\renewcommand{\thefootnote}{\arabic{footnote}}
% \maketitle

\begin{abstract}
Self-consistency (SC), a widely used decoding strategy for chain-of-thought reasoning, shows significant gains across various multi-step reasoning tasks but comes with a high cost due to multiple sampling with the preset size. Its variants, Adaptive self-consistency (ASC) and Early-stopping self-consistency (ESC), dynamically adjust the number of samples based on the posterior distribution of a set of pre-samples, reducing the cost of SC with minimal impact on performance. Both methods, however, do not exploit the prior information about question difficulty. 
It often results in unnecessary repeated sampling for easy questions that could be accurately answered with just one attempt, wasting resources. 
To tackle this problem, we propose Difficulty-Adaptive Self-Consistency (DSC), which leverages the difficulty information of batch queries from both prior and posterior perspectives to adaptively allocate inference resources, further reducing the overall cost of SC. To demonstrate the effectiveness of DSC, we conduct extensive experiments on three popular categories of reasoning tasks: arithmetic, commonsense and symbolic reasoning on six benchmarks. The empirical results show that DSC consistently surpasses the strong baseline ASC and ESC in terms of costs by a significant margin, while attaining comparable performances.\footnote{Our code and data have been released on \url{https://github.com/WangXinglin/DSC}.}

\end{abstract}
\section{Introduction}
Large language models (LLMs) have exhibited strong reasoning capabilities \citep{bubeck2023sparks}, especially with chain-of-thought (CoT) prompting \citep{wei2022chain}. Based on this, \citet{wang2022self} introduced a simple decoding strategy called self-consistency (SC) to further improve reasoning performance, leveraging the fact that challenging reasoning tasks typically require more reasoning paths to arrive at the correct answer. In contrast to the standard chain-of-thought prompting which only generates the greedy one, this method samples multiple reasoning paths according to a preset sample size, and then derives the final answer through majority-voting-based scheme. 
% COT,SC 从难度角度叙述 从unified角度去叙述。
% 窗口大小动态

\begin{figure}[t]
\begin{center}
\includegraphics[width=0.38\textwidth]{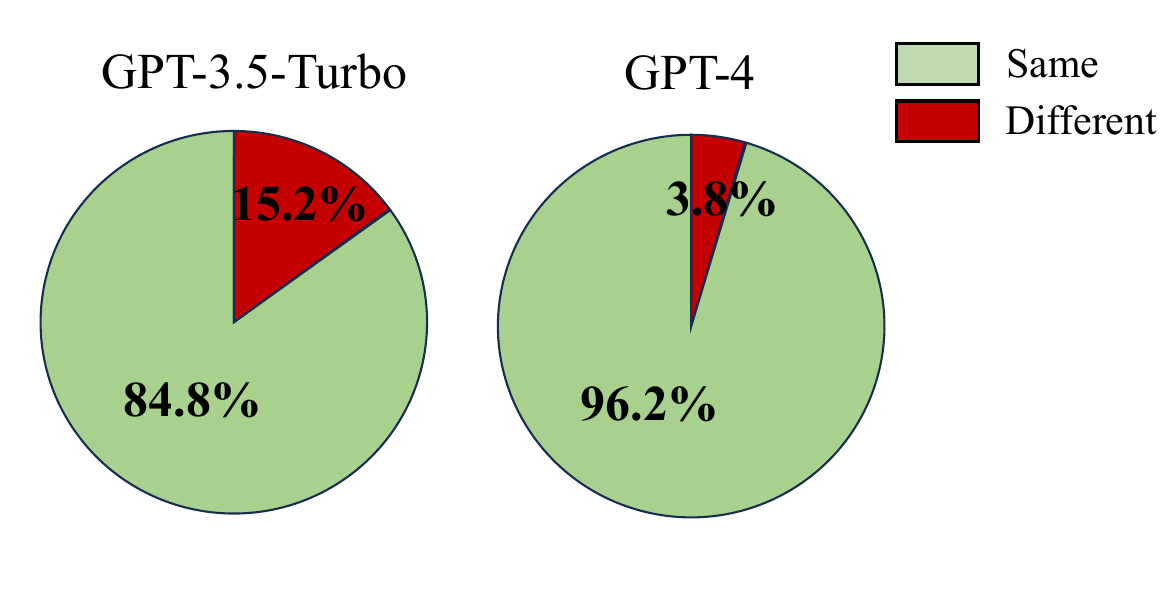}
\end{center}
\caption{The proportion of identical inference results between CoT and SC on GSM8K with GPT-3.5-Turbo and GPT-4. We set sample size of SC as 40.}
\label{fig:intro_analysis}
\end{figure}

% Efforts to reduce the cost of SC involve dynamically adjusting the number of samples based on the posterior distribution of a number of pre-samples. 
\begin{table}[t]
\centering
\small
\begin{tabular}{lcccc}
\toprule
\multicolumn{1}{c}{\multirow{2}{*}{Model}} & \multicolumn{2}{c}{Input} & \multicolumn{2}{c}{Output} \\ \cmidrule{2-5} 
\multicolumn{1}{c}{} & Token & Cost & Token & Cost \\ \toprule
GPT-3.5-Turbo & 846.3 & 0.0004 & 163.7 & 0.0002 \\
GPT-4 & 846.3 & 0.0254 & 142.1 & 0.0085 \\ \bottomrule
\end{tabular}
\caption{Average tokens and cost (\$) statistics of input and output for GPT-3.5-Turbo and GPT-4 on GSM8K. The cost is calculated according to \url{https://openai.com/pricing}. Given that the input for reasoning tasks usually involves several demonstrations (leading to lengthy contexts), the cost of input cannot be overlooked.}
\label{tb:intro_input_token}
\end{table}

\begin{figure*}[t]
\begin{center}
\includegraphics[width=0.79\textwidth]{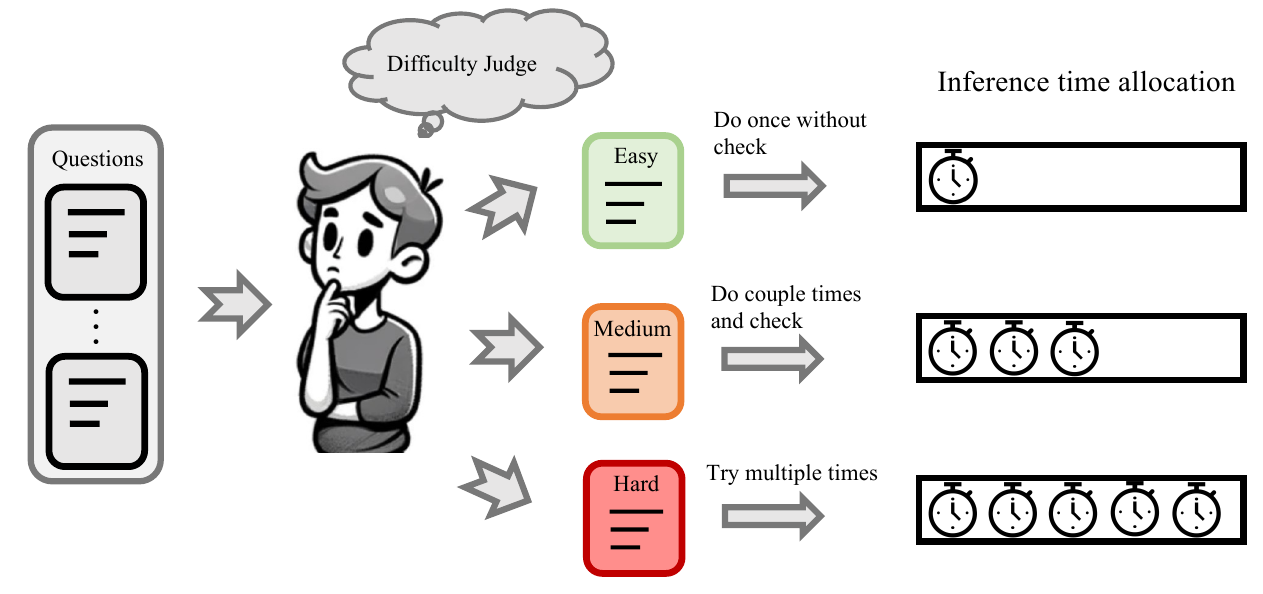}
\end{center}
\caption{Leveraging their prior knowledge, humans assess the difficulty level of a problem before solving it, and allocate appropriate time for its resolution based on the difficulty.}
\label{fig:intro_human}
\end{figure*}

Despite generally leading to improvements, SC introduces a significant overhead proportional to the number of sampled outputs. As LLMs continue to grow in size and complexity, the sampling time and computational costs associated with majority voting become increasingly challenging. 
Recently, some works seek to reduce the cost of SC by dynamically adjusting the number of samples based on the posterior distribution of pre-samples. 
ASC \citep{ASC} samples one by one, and stops sampling when the existing sample answers have established a clear majority as judged by a lightweight stopping criterion. Alternatively, ESC \citep{ESC} divides the large preset sample size into several sequential small windows, and stop sampling when answers within a window are all the same. 

% some works seek to estimate the difficulty of each question based on the posterior distribution of a number of pre-samples, thereby dynamically adjusting the number of samples and reducing the cost of SC.

%\footnote{We calculate cost according to \url{https://openai.com/pricing}} 

However, both approaches still suffer from the following two shortcomings:
% (before reaching the stopping criterion)
(1) Both require a certain amount of pre-sampling for all problems, which still results in redundant waste. As shown in Figure \ref{fig:intro_analysis}, there is a high overlap of inference results between SC and CoT, which suggests that only a small portion of questions (3.8\% for GPT-4) benefit from SC. Generating multiple samples for the remaining questions compared to CoT (only sampling once) results in a significant waste of costs. Although ASC and ESC somewhat reduce the ineffective cost of SC by decreasing average sample size, there remains redundant sampling for simple problems\footnote{At least 3 samples per problem for ASC and 4 for ESC.}.
% where the CoT inference matches that of SC  
(2) Multiple re-samples bring additional significant input costs. Both ASC and ESC focus solely on reducing the number of outputs, without considering the extra input costs brought by multiple re-samples (Table \ref{tb:intro_input_token}). Consequently, for problems requiring multiple sampling times, ASC and ESC will introduce substantial extra input costs, sometimes outweighing the savings achieved through output sampling reduction (Table \ref{tb:various max sampling size}).

\begin{figure*}[th]
\begin{center}
\includegraphics[width=0.94\textwidth]{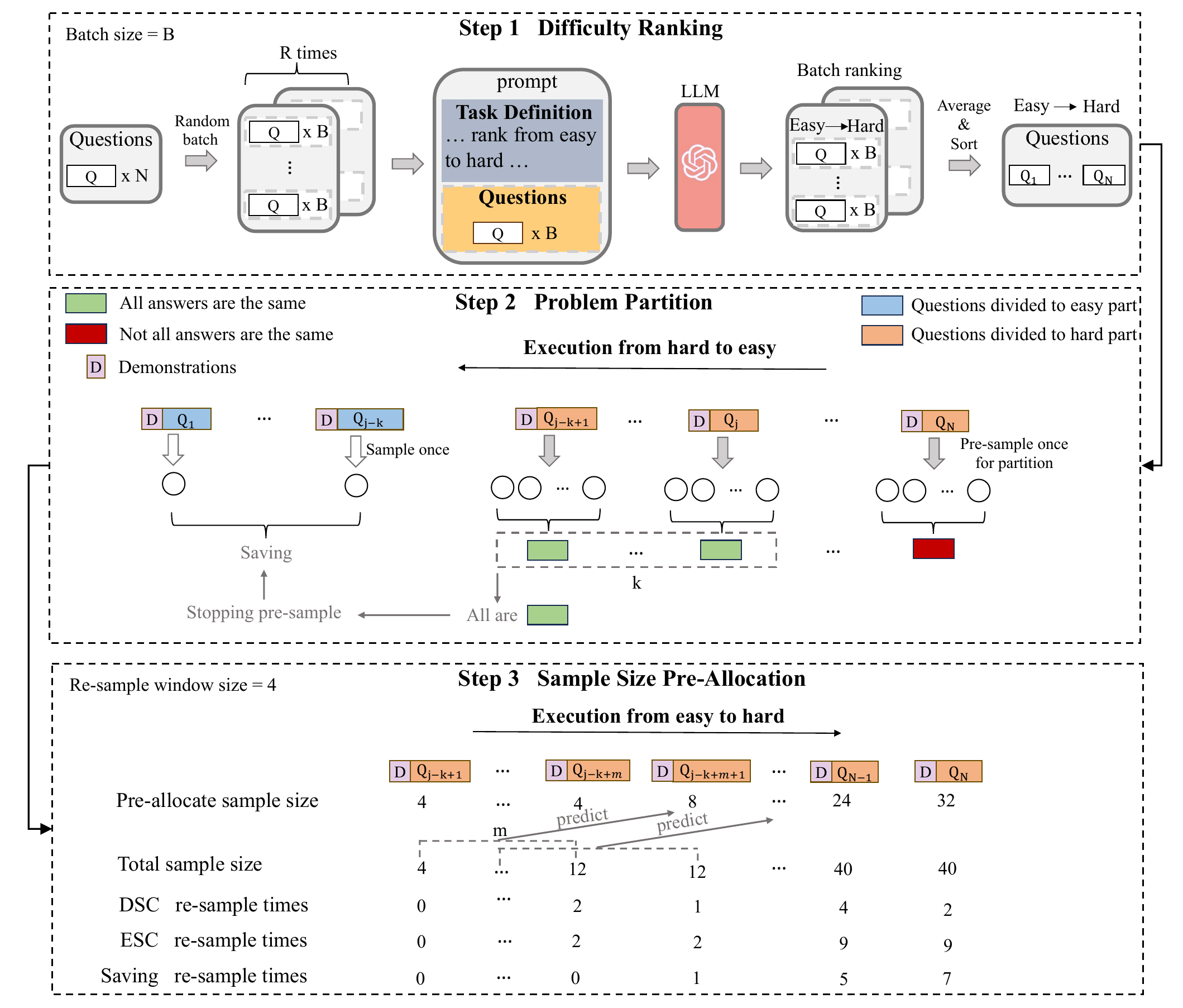}
\end{center}
\caption{Overall workflow of proposed Difficulty-Adaptive Self-Consistency. DSC first ranks problem difficulty using LLM itself (step 1), then partitions problems into easy and hard to save sampling cost for easy ones (step 2), and finally pre-allocates sample sizes to reduce resampling costs for hard problems (step 3).}
\label{fig:overview}
\end{figure*}

To alleviate these issues, we take inspiration from the strategies humans employ when solving reasoning problems within a limited total time. As illustrated in Figure \ref{fig:intro_human}, humans pre-assess the difficulty of the problems before reasoning, and adaptively allocating problem-solving time based on the accessed difficulty. 
%For simple problems, they would typically complete it once without check, allocating a minimal amount of time. However, for tasks of medium to high difficulty, they would experiment with various problem-solving methods and conduct repeated checks, thus allocating more time. 
% In light of this, we propose Difficulty-Adaptive Self-Consistency (DSC), a novel method which leverages the difficulty information accessed by LLM itself, together with the pre-sample distribution of a subset of given problems, to further reduce the inference computational cost of SC. 
In light of this, we propose Difficulty-Adaptive Self-Consistency (DSC), a novel method which leverages the difficulty information from both prior and posterior perspectives to further reduce the inference computational cost of SC. 
As illustrated in Figure \ref{fig:overview}, DSC consists of three steps. Firstly, we propose Difficulty Ranking algorithm which utilizes the LLM itself to rank the difficulty of the given problem set. While using LLM for ranking will incur additional costs, later experiments will demonstrate that the prior difficulty information gained from this process can lead to a greater reduction in overall inference costs. Based on this, we propose a Problem Partition strategy that divides the problem set into two parts, easy and hard, using the information regarding difficulty rankings. For problems belonging to easy part, a single CoT sampling is performed, which saves cost without sacrificing performance. Lastly, we propose Sample Size Pre-Allocation algorithm to predict the sample size needed for problems belonging to hard part, which reduce the number of re-samples and save the cost of inputs.

%  Although using LLM for difficulty ranking will introduce additional cost, our later experiments will show that prior difficulty gained can significantly reduce overall inference costs. 

We evaluate DSC on a wide range of arithmetic, commonsense and symbolic reasoning tasks over GPT-3.5-Turbo and GPT-4 models. The empirical results demonstrate that DSC outperforms the existing strong baselines ASC and ESC by a significant margin on six popular benchmarks, while attaining comparable performances.  Further experiments confirm the effectiveness of the proposed three-step algorithms: Difficulty Ranking, Problem Partition, and Sample Size Pre-Allocation.
%Firstly, it ranks the given question set based on difficulty ranking algorithm. Then, using Problem Partition strategy, it divides the problem set into two parts: easy and hard. For the easy part, only one COT sampling is performed to save cost. Finally, the proposed Sample Size Pre-Allocation algorithm is used to predict the sample size needed for the hard part of the problem, reducing the number of re-samples and thus saving the cost of the input part.

In summary, this work includes the following key contributions: 
\begin{itemize}
    \item We analyzed two common issues present in existing SC variants, ESC and ASC, designed for cost-efficient decoding.
    \item We proposed Difficulty-Adaptive Self-Consistency, a novel method consisting of three steps to fully utilize the difficulty information of given problems so as to reduce the inference computational cost.
    \item We conducted extensive experiments on six popular benchmarks and validated the effectiveness and generalizability of our proposed method.
    \item To the best of our knowledge, we are the first to propose dynamically allocating sampling resources based on query difficulty.
\end{itemize}

\section{Methodology}
The core idea behind DSC is to fully utilize the difficulty information of given problems, allocating computational resources based on their respective levels of difficulty.
The overall workflow of DSC is presented in Figure \ref{fig:overview}, including three steps: Difficulty Ranking, Problem Partition and Sample Size Pre-Allocation. 

\subsection{Difficulty Ranking}
As problems usually do not carry difficulty label themselves, we seek to utilize the powerful comparative and ranking capabilities of LLM to obtain the difficulty rank of the given problem set. Considering the limited context window size of LLM and its \textit{lost in the middle} \citep{liu-etal-2024-lost} issue in understanding long contexts, we randomly divide the problem set of $N$ into batches of size $B$ ($B<<N$) and let LLM rank the difficulty of the problems in each batch. Since a single random split only allows us to obtain the difficulty ranking of each problem within the corresponding batch, we perform $R$ times random batch splitting, and sort the average difficulty rankings of each problem in different batches to obtain the difficulty ranking of entire problem set. Algorithm \ref{alg:Difficulty ranking} illustrates the specific implementation of difficulty ranking. Specifically,  $D_{all}$ is a dictionary that stores the difficulty levels of all $N$ questions. Its keys represent the indices of the questions, and its values are lists used to store the difficulties of the questions across different batches. Merge indicates appending the difficulty of question $i$ in the current batch $r$ to $D_{all}[i]$ (List). Average refers to calculating the average value of $D_{all}[i]$, so as to obtain difficulty value of question i across the entire problem set. Given the additional costs of using LLM for ranking, we carefully design the ranking prompts\footnote{See Appendix \ref{prompt} for corresponding prompts.} to minimize these costs, and later experiments will prove that the prior difficulty information gained can significantly reduce overall inference costs.

\begin{algorithm}[t]
\small
    \caption{Difficulty Ranking.}\label{alg:Difficulty ranking}
        \begin{algorithmic}[1]
\Require Questions $Q^{1:N}$, LLM $\mathcal{M}$, 
        Ranking prompt $P$, \newline 
        random split rounds $R$, Batch size $B$
        
\Ensure Questions sorted by difficulty $\Bar{Q}^{1:N}$
% \State $C^{1:k} \leftarrow Kmeans(X^{1:N})$
% \State $L\leftarrow \varnothing$
\State $D_{all} \leftarrow \{i:[\ ]$ for $i$ $\in [1,N]\}$
\For{$r\gets 1, R$}
    \State Randomly divide $Q^{1:N}$ into batches $b_r^{1:L}$, $L=\lceil \frac{N}{B} \rceil$
    \For{$i\gets 1, N$}
        \State $D_{current} \leftarrow \varnothing$
        \For{$j\gets 1, L$}:
            \State $D_{current} \leftarrow D_{current}$.Append$(\mathcal{M}(P,b_r^j))$
        \EndFor
        \State $D_{all} \leftarrow D_{all}$.Merge$(D_{current})$
    \EndFor
\EndFor
\State $\Bar{Q}^{1:N} \leftarrow$Sort(Average$(D_{all}))$
\end{algorithmic}
\end{algorithm}

% $\Bar{q}^{1:N}$

\subsection{Problem Partition}
\label{sc:problem partition}
After obtaining the problem set with difficulty ranking, we aim to find which part of the problems only require LLM to perform CoT sampling once. A simple and intuitive idea is that when LLM is very confident about a number of continuous problems sorted by difficulty (the results of all pre-samples are the same), it can be inferred that the problems easier than these problems only need one CoT sampling. Guided by this, we design the Problem Partition algorithm, illustrated in Algorithm \ref{alg:Problem Partition}.
Specifically, we pre-sample the problem set sorted by difficulty from hard to easy, and store the entropy of pre-sampling results of each problem ($S_{current}$) in a list. We stop pre-sampling when the latest $k$ items in the list are all zero. The remaining problems without pre-sampling are then divided into the easy part, and a single CoT sampling is performed for each.

\begin{algorithm}[t]
\small
    \caption{Problem Partition.}\label{alg:Problem Partition}
        \begin{algorithmic}[1]
\Require Questions from \textbf{hard to easy} ${Q}^{1:N}$, LLM $\mathcal{M}$,\newline
Demonstrations $D$, Pre-sample size $p$, Judge window $k$ 
\Ensure Anchor point $A$, Easy part questions ${Q}_{Easy}$,\newline 
Hard part questions ${Q}_{Hard}$
% \State $C^{1:k} \leftarrow Kmeans(X^{1:N})$
% \State $L\leftarrow \varnothing$
\State $E_{all} \leftarrow [\ ]$; $a \gets 1$
\For{$i\gets 1, N$}
    \State $S_{current} \leftarrow \mathcal{M}({Q}^{i},D,p)$
    \State $E_{all} \leftarrow E_{all}$.Append$(Entropy(S_{current}))$
    \If{$i>k$ and $Sum(E_{all}[i-k:i-1]) = 0$}
        \State $A \gets i$
        \State \textbf{Break}
    \EndIf
\EndFor

\State ${Q}_{Hard} \gets {Q}^{1:A} $; ${Q}_{Easy} \gets {Q}^{A+1:N} $
\end{algorithmic}
\end{algorithm}

\subsection{Sample Size Pre-Allocation}
After performing CoT sampling on the easy part problems, we seek to predict and allocate the sample size for the problems belonging to hard part, so as to mitigate the substantial cost brought by multiple re-samples. Considering that the required sample size for each problem should be similar to those needed for problems of comparable difficulty, we predict the sample size of the current problem based on the total sample size of the nearest $m$ easier\footnote{As the pre-allocated sample size could be larger than the required sample size (causing higher output costs), we use the sample size of easier rather than harder neighboring problems for prediction.} problems. Algorithm \ref{alg:Sample Size Pre-Allocation} shows the workflow of Sample Size Pre-Allocation. Specifically, we sample questions belonging to the hard part from easy to difficult. For the current question, we predict its pre-allocation sample size $PA$ based on the average total sample size of its previous $m$ questions. Then, we judge the distribution of the current samples based on the stopping criteria $C$\footnote{Following ASC, we use Dirichlet Stopping Criteria as default. See Appendix \ref{sec: background} for more details.}. When the criteria is not met, we re-sample based on the expansion window $e$, until the sampling distribution meets the criteria for stopping sampling or the number of samples reaches the max sample size $L$. After sampling for the current question ends, we add its samples and total sample size to $S_{all}$ and $N_{all}$ lists respectively, in order to pre-allocate the sample size for the next question.

\begin{algorithm}[t]
\small
    \caption{Sample Size Pre-Allocation.}\label{alg:Sample Size Pre-Allocation}
        \begin{algorithmic}[1]
\Require Hard part questions from \textbf{easy to hard} ${Q}^{1:A}$, \newline LLM $\mathcal{M}$,
Demonstrations $D$, Stopping Criteria $C$, \newline 
Max sample size $L$, Extend window size $e$, Prediction window size $m$
\Ensure Sampling of ${Q}^{1:A}$ based on pre-allocation $S_{all}$
\State $S_{all}\leftarrow \{i:[\ ]$ for $i$ $\in [1,A]\}$; $N_{all}\leftarrow [\ ]$
\For{$i\gets 1, A$}
    \State $S_{current} \leftarrow \varnothing$; $PA \leftarrow 0$; $N_{current} \leftarrow 0$
    \If{$i>m$}
        \State $PA \gets Average(N_{all}[i-m:i-1])$
        \State $S_{current} \leftarrow S_{current}$.Append$(\mathcal{M}({Q}^{i},D,PA))$
        \State $N_{current} \leftarrow N_{current}+PA$
    \EndIf
    
    \While{$N_{current} < L$ and NOT $C(S_{current})$}
        \State $S_{current} \leftarrow S_{current}$.Append$(\mathcal{M}({Q}^{i},D,e))$
        \State $N_{current} \leftarrow N_{current}+e$
    \EndWhile
    
    \State $S_{all} \leftarrow S_{all}$.Merge$(S_{current})$
    \State $N_{all} \leftarrow N_{all}$.Append$(N_{current})$
    
\EndFor
\end{algorithmic}
\end{algorithm}

% As shown in Figure \ref{fig:overview}, the workflow of DSC consists of three steps. The first step is to rank the difficulty of the given problem set through the Difficulty Ranking algorithm and obtain relative difficulty information. The second step, based on the obtained relative difficulty information and combined with the distribution of a small number of pre-samples of some problems, estimates the absolute difficulty of the given problems through the problem partition algorithm. This allows the problem set to be divided into easy and hard parts based on absolute difficulty, thereby allocating reasonable computational resources for simple problems. The third step is to estimate the computational resources needed for the hard part of problems through the Sample Size Pre-Allocation algorithm, thereby alleviating the large amount of additional input cost brought by multiple resamples.
\section{Experiments}

\subsection{Experimental Setup}
\subsubsection{Benchmarks}
We evaluate the proposed DSC on six benchmark datasets from three categories of reasoning tasks:
For arithmetic reasoning, we consider MATH \citep{MATH} and GSM8K \citep{GSM8K}. MultiArith \citep{multi}, SVAMP \citep{svamp}, AddSub \citep{AddSub} and ASDiv \citep{asdiv} are not chosen in this paper because they are relatively simple.
For commonsense reasoning, CommonsenseQA \citep{CSQA} and StrategyQA \citep{SQA} are used. For symbolic reasoning, we use Last Letter Concatenation and Coin Flip from \citet{COT}. The data version is from \citet{ZeroCOT}.

\subsubsection{Baselines}
We compare DSC to the following self-consistency methods:
(1) SC \citep{SC} is the standard self-consistency which samples multiple reasoning paths and then derives the final answer through majority-voting; (2) ASC \citep{ASC} samples one by one, and stops sampling when the existing samples meet a designed stopping criteria, which measures the LLM's confidence in its  current samples; (3) ESC \citep{ESC} proposes using small window detection to save cost, which divides the large preset sample size into several sequential small windows, and stops sampling when answers within a window are all the same. Specifically, ASC and ESC are two strong baselines for cost-efficient self-consistency methods, we reproduce both methods according to their original implementation.

\subsubsection{Implementation details}
We perform experiments using two powerful LLMs: GPT-4 \citep{GPT4} and GPT-3.5-Turbo\footnote{We use the "2023-05-15" version of API for both.}. We calculate the cost based on the price of the API we use. All experiments are conducted in the few-shot setting without training or fine-tuning the language models. To ensure a fair comparison, we use the same prompts as \citet{COT}. 
% Details on the prompts used are given in Appendix.
Following \citet{ESC}, The sampling temperature $T$ for MATH is 0.5 while for other datasets is 0.7. For difficulty ranking, we use CoT sampling. We set the default parameters as follows: batch size $B$ as 8 and random split rounds $R$ as 5 for Difficulty Ranking; pre-sample size $p$ as 4 and judge window size $k$ as 32 for Problem Partition; extend window size $e$ as 4 and prediction window size $m$ as 16 for Sample Size Pre-Allocation; max sample size $L$ as 40 for all baselines. All experiments are repeated 100 times and the average performance is reported. Unless otherwise specified, the reported cost of DSC includes the cost brought by all three sub-steps.

\begin{table*}[t]
\renewcommand{\arraystretch}{0.95}
\setlength{\tabcolsep}{3.6pt}
\centering
\small
\begin{tabular}{llcccccccccccc}
\toprule
\multirow{2}{*}{Model} & \multicolumn{1}{c}{\multirow{2}{*}{Method}} & \multicolumn{2}{c}{MATH} & \multicolumn{2}{c}{GSM8K} & \multicolumn{2}{c}{CSQA} & \multicolumn{2}{c}{SQA} & \multicolumn{2}{c}{Letter} & \multicolumn{2}{c}{Coinflip} \\ \cmidrule{3-14} 
 & \multicolumn{1}{c}{} & Cost & Acc & Cost & Acc & Cost & Acc & Cost & Acc & Cost & Acc & Cost & Acc \\ \toprule
\multirow{4}{*}{GPT-4} & SC & 0.4220 & 58.52 & 0.3509 & 95.81 & 0.1280 & 85.82 & 0.1493 & 81.89 & 0.1691 & 94.51 & 0.1528 & 100 \\
 & ASC & 0.5726 & 58.48 & 0.1712 & 95.79 & 0.1639 & 85.82 & 0.1323 & 81.89 & 0.0602 & 94.50 & 0.0657 & 100 \\
 & ESC & 0.4062 & 58.49 & 0.1014 & 95.80 & 0.0767 & 85.81 & 0.0647 & 81.89 & 0.0375 & 94.51 & 0.0338 & 100 \\
 & DSC & \textbf{0.3142} & 58.51 & \textbf{0.0699} & 95.79 & \textbf{0.0598} & 85.81 & \textbf{0.0516} & 81.88 & \textbf{0.0259} & 94.50 & \textbf{0.0264} & 100 \\ \midrule
\multirow{4}{*}{GPT-3.5-Turbo} & SC & 0.0181 & 49.43 & 0.0094 & 83.22 & 0.0030 & 76.80 & 0.0043 & 71.47 & 0.0043 & 80.44 & 0.0037 & 76.60 \\
 & ASC & 0.0193 & 49.39 & 0.0064 & 83.18 & 0.0034 & 76.78 & 0.0032 & 71.47 & 0.0018 & 80.43 & 0.0034 & 76.59 \\
 & ESC & 0.0172 & 49.41 & 0.0048 & 83.19 & 0.0017 & 76.79 & 0.0019 & 71.47 & 0.0014 & 80.43 & 0.0020 & 76.59 \\
 & DSC & \textbf{0.0148} & 49.42 & \textbf{0.0036} & 83.19 & \textbf{0.0012} & 76.79 & \textbf{0.0015} & 71.46 & \textbf{0.0011} & 80.42 & \textbf{0.0016} & 76.60 \\ \bottomrule
\end{tabular}
\caption{Accuracy (\%) and cost (\$) across six reasoning benchmarks. The best performance of cost is highlighted in \textbf{bold}.}
\label{tb:main result}
\end{table*}

\definecolor{lightgreen}{RGB}{48,128, 20} % 自定义绿色
\definecolor{lightred}{RGB}{220, 54, 20} % 自定义绿色

\begin{table*}[t]
\renewcommand{\arraystretch}{0.95}
\setlength{\tabcolsep}{4.8pt}
\centering
\small
\begin{tabular}{llcccccccccc}
\toprule
\multirow{2}{*}{Model} & \multirow{2}{*}{Method} & \multicolumn{2}{c}{16} & \multicolumn{2}{c}{24} & \multicolumn{2}{c}{32} & \multicolumn{2}{c}{40} & \multicolumn{2}{c}{48} \\ \cmidrule{3-12} 
 &  & Cost & Acc & Cost & Acc & Cost & Acc & Cost & Acc & Cost & Acc \\ \toprule
\multirow{4}{*}{GPT-4} & SC & 0.1791 -  & 57.34 & 0.2602  -  & 57.95 & 0.3411  -  & 58.27 & 0.4220  -  & 58.52 & 0.5030  -  & 58.67 \\
 & ASC & 0.3038 $\downarrow$ & 57.33 & 0.4038  $\downarrow$ & 57.93 
& 0.4923 $\downarrow$ & 58.23 & 0.5726 $\downarrow$ & 58.48 & 0.6470 $\downarrow$ & 58.62 \\
 & ESC & 0.1870 $\downarrow$ & 57.34 & 0.2631 $\downarrow$ & 57.94 & 0.3361 $\uparrow$ & 58.25 & 0.4062 $\uparrow$ & 58.49 & 0.4723 $\uparrow$ & 58.64 \\
 & DSC & \textbf{0.1659} $\uparrow$ & 57.34 & \textbf{0.2192} $\uparrow$ & 57.94 & \textbf{0.2680} $\uparrow$ & 58.25 & \textbf{0.3142} $\uparrow$ & 58.51 & \textbf{0.3585} $\uparrow$ & 58.66 \\ \midrule
\multirow{4}{*}{GPT-3.5-Turbo} & SC & 0.0074 -  & 47.98 & 0.0110 -  & 48.71 & 0.0145 -  & 49.13 & 0.0181 -  & 49.43 & 0.0216 -  & 49.67 \\
 & ASC & 0.0095 $\downarrow$ & 47.97 & 0.0131 $\downarrow$ & 48.69 & 0.0163 $\downarrow$ & 49.10 & 0.0193 $\downarrow$ & 49.39 & 0.0220 $\downarrow$ & 49.62 \\
 & ESC & 0.0075 $\downarrow$ & 47.98 & 0.0108 $\uparrow$ & 48.70 & 0.0140 $\uparrow$ & 49.12 & 0.0172 $\uparrow$ & 49.41 & 0.0203 $\uparrow$ & 49.65 \\
 & DSC & \textbf{0.0069} $\uparrow$ & 47.98 & \textbf{0.0097} $\uparrow$ & 48.71 & \textbf{0.0123} $\uparrow$ & 49.13 & \textbf{0.0148} $\uparrow$ & 49.42 & \textbf{0.0171} $\uparrow$ & 49.66 \\ \bottomrule
\end{tabular}
\caption{Reasoning accuracy (\%) and corresponding cost (\$) with various max sampling size on MATH. We mark the performance of cost poorer than SC with $\downarrow$ and that better than SC with $\uparrow$. The best performance of cost is highlighted in \textbf{bold}.}
\label{tb:various max sampling size}
\end{table*}

\begin{table}[t]
\renewcommand{\arraystretch}{0.92}
\setlength{\tabcolsep}{5pt}
\small
\centering
\begin{tabular}{lcccc}
\toprule
Method & Input tokens & Output tokens & Cost & Acc \\ \toprule
SC & 846 & 6045 & 0.3881 & 73.19 \\
ASC & 15982 & 3265 & 0.6754 & 73.14 \\
ESC & 4458 & 4415 & 0.3986 & 73.18 \\
DSC & 1617 & 3857 & 0.2799 & 73.18 \\ \bottomrule
\end{tabular}
\caption{Accuracy (\%) and cost (\$) on GSM8K with small open-source language model Mistral-7B-Instruct-v0.3.}
\label{tb:small open-source models}
\end{table}

\subsection{Main Results}

\paragraph{DSC significantly reduces costs while barely affecting performance.}
Table~\ref{tb:main result} summarizes the cost and accuracy of SC, ASC, ESC, and proposed DSC for each dataset. We show that DSC consistently outperforms all baselines on cost by a significant margin across all datasets, while barely affecting performance. Specifically, DSC reduces the cost on GPT-4 by an average of 65.29\% and on GPT-3.5-Turbo by 56.04\% compared to SC. In comparison to the strongest baseline method ESC, DSC reduces the cost on GPT-4 by an average of 24.81\% and on GPT-3.5-Turbo by 21.86\%, which demonstrates the effectiveness of DSC. Furthermore, the above results show that the prior difficulty information gained from difficulty ranking can help greatly lower overall inference costs, even though it first introduces some additional costs.

\paragraph{DSC is scalable across various max sampling size.}
We conduct experiments with various max sampling size to validate the scalability of ESC. Table~\ref{tb:various max sampling size} shows the performance across different maximum sampling sizes. First we can see the performance of SC continuously improves as max sampling size $L$ increases, which is consistent with the results in \citep{SC}. On this basis, ESC can significantly save costs while maintaining performance for different $L$. On the contrary, due to the substantial additional input cost brought by multiple re-samples, ASC and ESC result in higher cost than SC when the cost savings of output tokens are limited.

\paragraph{DSC Maintains Generalizability on Small Open-source Language Model.}
To further explore the effectiveness of DSC on small open-source models, we conduct experiments on the open-source model Mistral-7B-Instruct-v0.3 \citep{mistral} using GSM8K dataset. As shown in Table \ref{tb:small open-source models}, we convert the token cost into price according to that of GPT-4 for a simpler comparison and keep the other settings completely consistent with the main experiment\footnote{Please note that for DSC, the cost of all three sub-steps are included.}. The experimental results indicate that DSC has the potential to work effectively on smaller models.

% \usepackage{multirow}

\begin{table*}[t]
\setlength{\tabcolsep}{3.6pt}
\renewcommand{\arraystretch}{0.90}
\centering
\small
\begin{tabular}{llcccccccc}
\toprule
\multirow{2}{*}{Model} & \multirow{2}{*}{Metric} & \multicolumn{1}{l}{\multirow{2}{*}{Geometry}} & \multicolumn{1}{l}{Counting \&} & \multicolumn{1}{l}{\multirow{2}{*}{Prealgebra}} & \multicolumn{1}{l}{Intermediate} & \multicolumn{1}{l}{Number} & \multicolumn{1}{l}{\multirow{2}{*}{Precalculus}} & \multicolumn{1}{l}{\multirow{2}{*}{Algebra}} & \multicolumn{1}{l}{\multirow{2}{*}{Avg}} \\
 &  & \multicolumn{1}{l}{} & Probability & \multicolumn{1}{l}{} & Algebra & Theory & \multicolumn{1}{l}{} & \multicolumn{1}{l}{} & \multicolumn{1}{l}{} \\ \toprule
\multirow{3}{*}{GPT-4} & Spearman & 0.540 & 0.578 & 0.565 & 0.654 & 0.654 & 0.684 & 0.691 & 0.624 \\
 & Pearson & 0.535 & 0.585 & 0.567 & 0.660 & 0.654 & 0.691 & 0.695 & 0.627 \\
 & Kendall & 0.417 & 0.448 & 0.440 & 0.519 & 0.520 & 0.547 & 0.547 & 0.491 \\ \midrule
\multirow{3}{*}{GPT-3.5-Turbo} & Spearman & 0.415 & 0.307 & 0.362 & 0.484 & 0.445 & 0.497 & 0.573 & 0.440 \\
 & Pearson & 0.415 & 0.314 & 0.370 & 0.495 & 0.452 & 0.514 & 0.578 & 0.448 \\
 & Kendall & 0.318 & 0.233 & 0.279 & 0.375 & 0.344 & 0.383 & 0.446 & 0.340 \\ \bottomrule
\end{tabular}
\caption{Performance of Difficulty Ranking on MATH dataset.}
\label{tb:rank result}
\end{table*}

% \usepackage{multirow}
\begin{table*}[t]
\renewcommand{\arraystretch}{0.92}
\setlength{\tabcolsep}{3.42pt}
\centering
\small
\begin{tabular}{llcccccccccccc}
\toprule
\multirow{2}{*}{Model} & \multicolumn{1}{c}{\multirow{2}{*}{Method}} & \multicolumn{2}{c}{MATH} & \multicolumn{2}{c}{GSM8K} & \multicolumn{2}{c}{CSQA} & \multicolumn{2}{c}{SQA} & \multicolumn{2}{c}{Letter} & \multicolumn{2}{c}{Coinflip} \\ \cmidrule{3-14} 
 & \multicolumn{1}{c}{} & Input & Output & Input & Output & Input & Output & Input & Output & Input & Output & Input & Output \\ \toprule
\multirow{6}{*}{GPT-4} & SC & 576 & 6746 & 846 & 5426 & 726 & 1770 & 611 & 2184 & 302 & 2667 & 428 & 2332 \\
 & ASC & 11100 & 3994 & 4266 & 719 & 4805 & 327 & 3726 & 351 & 1397 & 308 & 1722 & 234 \\
 & ESC & 3574 & 4983 & 1423 & 982 & 1644 & 455 & 1228 & 465 & 454 & 401 & 540 & 294 \\ \cmidrule{2-14} 
 & DSC & 834 & 4606 & 975 & 495 & 978 & 271 & 822 & 310 & 314 & 113 & 428 & 58 \\
 & w/o step 2 & 847 & 4701 & 1035 & 755 & 1081 & 370 & 873 & 381 & 353 & 313 & 431 & 235 \\
 & w/o step 3 & 2896 & 4142 & 1033 & 479 & 1068 & 253 & 903 & 290 & 328 & 108 & 428 & 58 \\ \midrule
\multirow{6}{*}{GPT-3.5-Turbo} & SC & 576 & 11851 & 846 & 6005 & 726 & 1749 & 611 & 2644 & 302 & 2735 & 428 & 2316 \\
 & ASC & 13421 & 8380 & 8075 & 1577 & 5633 & 365 & 4698 & 535 & 2150 & 487 & 4932 & 624 \\
 & ESC & 4223 & 10045 & 2889 & 2273 & 1946 & 506 & 1599 & 731 & 742 & 673 & 1613 & 819 \\ \cmidrule{2-14} 
 & DSC & 793 & 9456 & 1410 & 1809 & 1140 & 376 & 985 & 591 & 470 & 535 & 704 & 736 \\
 & w/o step 2 & 799 & 9610 & 1426 & 1879 & 1172 & 401 & 997 & 605 & 482 & 584 & 707 & 761 \\
 & w/o step 3 & 3485 & 8634 & 2144 & 1679 & 1467 & 362 & 1238 & 560 & 568 & 513 & 1290 & 648 \\ \bottomrule
\end{tabular}
\caption{Ablation study of Problem Partition (step 2) and Sample Size Pre-Allocation (step 3). We count input and output tokens across six reasoning benchmarks. To simplify the comparison, the tokens produced by Difficulty Ranking are not counted here.}
\label{tb:step2 and step3}
\end{table*}
% input and output 

\begin{table}[t]
\setlength{\tabcolsep}{2.1pt}
\small
\centering
\begin{tabular}{lcccccc}
\hline
Task & MATH & GSM8K & CSQA & SQA & Letter & Coinflip \\ \hline
Ranking & 403 & 364 & 261 & 130 & 161 & 308 \\
Reasoning & 576 & 846 & 726 & 611 & 302 & 428 \\ \hline
\end{tabular}
\caption{Comparison of average input tokens for each question brought by Difficulty Ranking and few-shot reasoning across six reasoning datasets.}
\label{tb:rank cost}
\end{table}
% \usepackage{multirow}

\subsection{Analysis of Three Sub-steps}
To further validate the effectiveness of proposed three sub-steps, including Difficulty Ranking, Problem Partition, and Sample Size Pre-Allocation, we conduct experimental analyses on each of them respectively.
\subsubsection{Difficulty Ranking}
\label{subsub:difficulty ranking}
% To further analyze the cost and performance of proposed Difficulty Ranking algorithm, we compare the tokens produced by Difficulty Ranking and single few-shot reasoning across all six benchmark, and conduct experiments on MATH dataset, which has manually annotated difficulty labels.

\paragraph{Difficulty Ranking exhibits a good correlation with humans.}
As shown in Table \ref{tb:rank result}, we calculate Spearman, Pearson, and Kendall correlation coefficients on seven subsets of MATH benchmark. Overall, GPT-4 demonstrates a high consistency with human labels, while the weaker GPT-3.5-Turbo shows a moderate consistency. Specifically, both GPT-4 and GPT-3.5-Turbo rank weakly on Geometry difficulty, which may be due to the fact that text LLMs cannot intuitively assess the difficulty of geometry problems through visual information like humans can.

\paragraph{Difficulty Ranking produces an acceptable cost.}
As we instruct LLM to produce extremely concise outputs for Difficulty Ranking, the cost of output tokens brought by Difficulty ranking is very low\footnote{With an average consumption of 15 tokens for each question.}.
Table \ref{tb:rank cost} gives comparison of input tokens brought by Difficulty Ranking and few-shot reasoning across six datasets. We show that the average input cost for each problem's difficulty ranking is less than the average single input cost for its few-shot reasoning. Considering that ESC and ASC require multiple re-samples for reasoning (multiple few-shot inputs), this additional input cost of Difficulty Ranking is totally acceptable.

% \paragraph{Mixing types of questions lowers the effectiveness of Difficulty Ranking.}
% To explore whether Difficulty Ranking can effectively rank problems of different types in terms of difficulty, we compare the performance of Difficulty Ranking under mixed and unmixed problem types on MATH, as shown in Table \ref{tb:mix type}. Specifically, for unmixed, we perform Difficulty Ranking on seven subsets separately; for mixed, we directly execute Difficulty Ranking on random shuffled MATH dataset, and calculate the correlation on the entire dataset for both. The results indicate that mixing different types of questions for LLM to rank can lead to a decrease in performance. This might be due to the challenge for LLM in adhering to the same assessment standards for various types of questions.\footnote{It could also lead to inaccurate ranking for humans.}

% \usepackage{multirow}

\subsubsection{Problem Partition}
\paragraph{Problem Partition proves to be effective across all datasets.}
As shown in Table \ref{tb:step2 and step3}, we conduct an ablation study on Problem Partition (step 2). We show that the removal of Problem Partition results in an increase in both input and output tokens of DSC across all six datasets,  which validates the effectiveness and generalizability of Problem Partition. Furthermore, we find that the removal of Problem Partition has a small impact on output tokens when it comes to MATH dataset, which could be due to the fact that for datasets with overall higher difficulty, the easy part problem (only require one CoT sample) for LLM is fewer.

\begin{figure}[t]
\begin{center}
\includegraphics[width=0.48\textwidth]{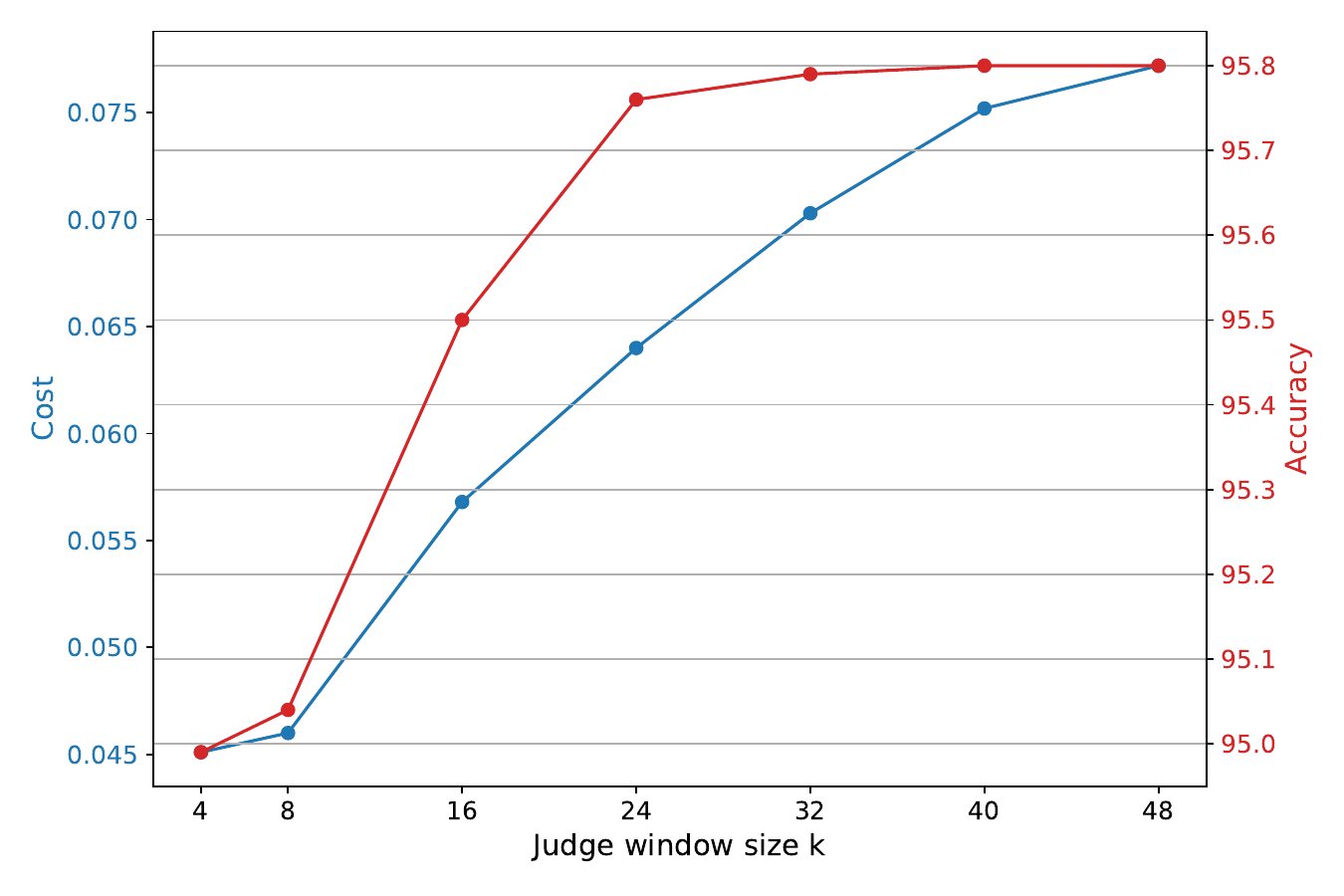}
\end{center}
\caption{Cost (\$) and accuracy (\%) of DSC under different judge window size $k$ with GPT-4 on GSM8K.}
\label{fig:accuracy of judge window size}
\end{figure}

\paragraph{Judge window size is essential for the accuracy of DSC.}
Regarding the intuitive idea proposed in Section \ref{sc:problem partition}, a straightforward question is how large the judge window size $k$ needs to be to ensure the accuracy on simpler problems with single CoT sampling. To answer this, we conduct an experiment of DSC under different judge window size $k$ with GPT-4 on GSM8K. As shown in Figure \ref{fig:accuracy of judge window size}, the accuracy increases with the rise of $k$ until it reaches 32. Meanwhile, the cost of DSC also increases with the enlargement of $k$, as the easy part divided by the problem partition decreases with its increase. Therefore, it is a good choice to set a smaller judge window size while ensuring accuracy.\footnote{We select 32 as the default value for $k$.}

\subsubsection{Sample Size Pre-Allocation}

\paragraph{Sample Size Pre-Allocation significantly alleviates the substantial input token costs brought by multiple re-samples.}
We conduct an ablation study of proposed Sample Size Pre-Allocation (step 3) to validate its effectiveness. As shown in Table \ref{tb:step2 and step3}, with the removal of Sample Size Pre-Allocation, the count of input tokens on DSC rose on all datasets on GPT-3.5-Turbo and GPT-4 (with the exception of GPT-4 on Coinflip), validating the effectiveness and generalizability of Sample Size Pre-Allocation. Moreover, we notice that the input tokens of ASC and ESC far exceed that of SC, which once again confirms that multiple re-samples will bring a large amount of additional overhead. Meanwhile, DSC significantly reduced the number of input tokens through Pre-Allocation, keeping it comparable to SC. In addition, we find that Sample Size Pre-Allocation leads to a slight increase in output tokens (due to the predicted sample size of some questions exceeding the actual requirement). Overall, we believe it is a good choice to trade a small number of output tokens for a significant saving in input tokens.

\begin{figure}[t]
\begin{center}
\includegraphics[width=0.48\textwidth]{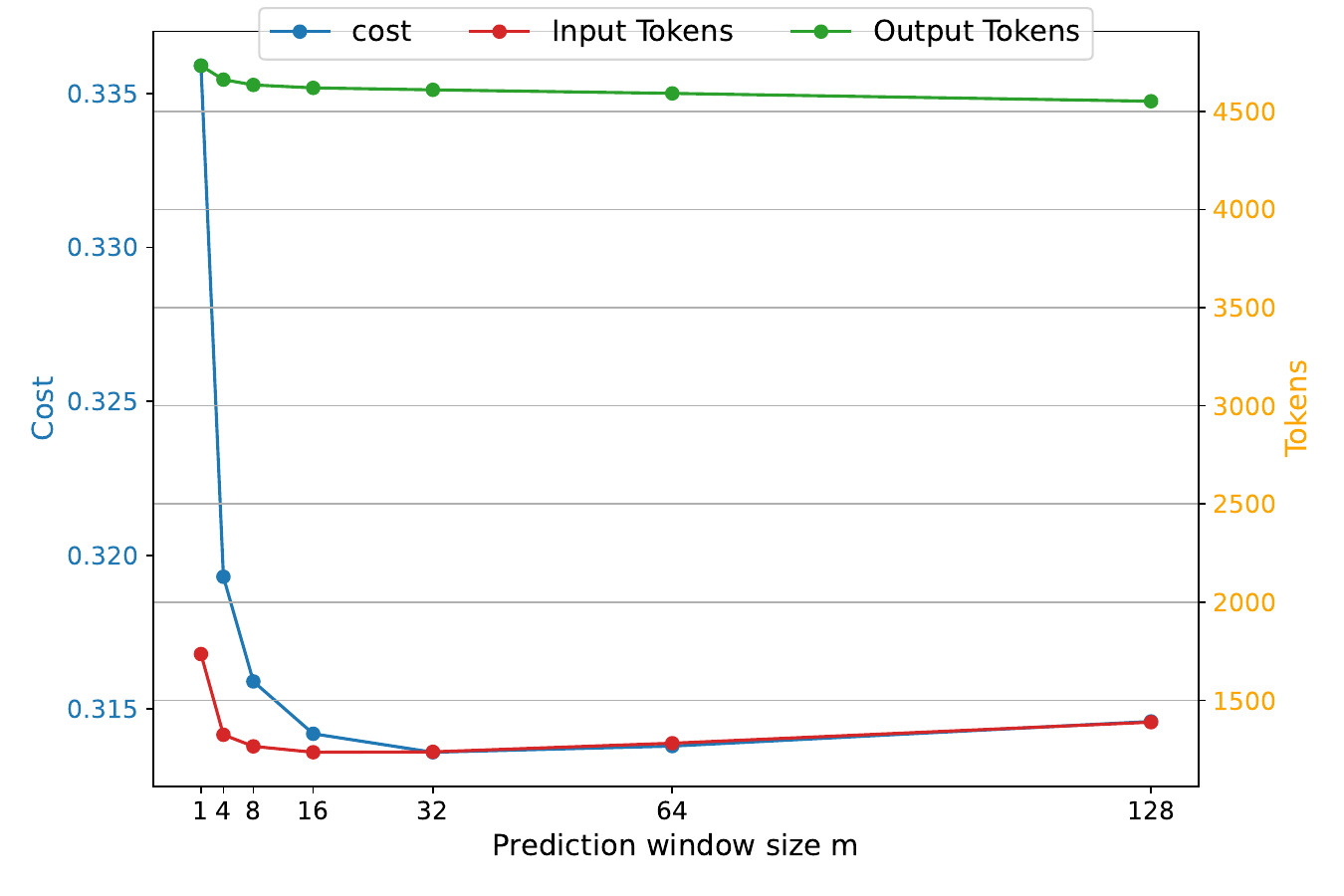}
\end{center}
\caption{Tokens of input and output and the corresponding cost (\$) of DSC under different prediction window size $m$ with GPT-4 on MATH.}
\label{fig:tokens of prediction window size}
\end{figure}

\paragraph{Prediction window size is the key to trade output tokens for more savings in input tokens.}
Given that Sample Size Pre-Allocation can trade output tokens for savings in input tokens, a natural question is when the max total savings can be achieved. As shown in Figure \ref{fig:tokens of prediction window size}, we count the tokens of input and output and the corresponding cost of DSC under different prediction window size $m$. The experimental results show that the cost of DSC decreases rapidly first and then slighly increases with the increase of $m$, and the input tokens also follow a similar trend. As $m$ rises, simpler questions predict larger sample sizes for the current query. However, when $m$ becomes sufficiently large, the predicted sample size falls below the actual value, leading to more re-sampling. In contrast, output tokens gradually decrease because, as $m$ increases, the predicted sample size continues to decline, reducing redundancy in output token sampling. In summary, it is suggested to select $m$ within a relatively moderate range (e.g. from 8 to 32).

% The experimental results show that the cost of DSC decreases rapidly first and then slighly increases with the increase of $m$, and the input tokens also follow a similar trend. This is because as $m$ increases, the sample size of the current question is predicted by more simpler questions. When the value of $m$ is large enough, the predicted sample size will gradually be less than its actual value, leading to the need for more re-samples. In contrast, the tokens of the output part gradually decrease, which is due to as m increases, the predicted sample size will continue to decrease, leading to a gradual decrease in the situation where the predicted sample size is greater than its actual need, saving the output tokens from redundant sampling.

\section{Related Work}

\paragraph{Chain-of-thought Reasoning}
Chain-of-thought prompting has been proven to be an effective method of solving complex reasoning problems \citep{COT}. By following the pattern of gradually solving sub-problems, few-shot CoT \citep{Specialize} are capable of stimulating LLM reasoning abilities. On this basis, Least-to-most prompting \citep{LeastToMost} suggest explicitly splitting the problem and solving them step by step. \citet{PHP} reach the final answer by iteratively generating answers and using the previously generated answers as context hints.

\paragraph{Self-Consistency}

Self-consistency \citep{SC} refers to a simple decoding strategy for further improving reasoning performance, leveraging the fact that complex reasoning tasks typically allow for more than one correct reasoning path. \citet{li2024turning} assign appropriate weights for answer aggregation to achieve adaptive self-consistency. \citet{open} and \citet{FSC} extend it for open-ended generation tasks like code generation and text summarization. However, they require multiple sampling with the pre-set size, which will incur much more computation cost. To achieve cost-efficient self-consistency, \citet{ASC} introduce an adaptive stopping criterion based on the amount of agreement between the samples so far. \citet{ESC} divide the large preset sample size into several sequential small windows, and stop sampling when answers within a window are all the same.

\paragraph{Difficulty-adaptive Test-Time Scaling}

Test-time scaling \citep{wu2024empirical, snell2024scaling, brown2024large} has been widely explored as an effective approach to improving model performance. However, conventional test-time scaling applies a uniform amount of computation to all queries, regardless of their difficulty, resulting in unnecessary computational overhead. To mitigate this inefficiency, recent work has proposed difficulty-adaptive scaling strategies that dynamically allocate computational resources based on query difficulty. \citet{snell2024scaling} validate the feasibility of optimizing test-time scaling compute based on query difficulty. To address excessive refinement in self-refinement, \citet{chen2024magicore} classify problems as easy or hard, solving easy ones with coarse aggregation and hard ones with fine-grained, iterative multi-agent refinement.
\citet{manvi2024adaptive} introduce capability-aware and mid-generation self-evaluations to enable difficulty-aware adaptive test-time computation by dynamically allocating compute based on query difficulty and model capability. 
Similarly, \citet{damani2024learning} train a lightweight model to estimate query difficulty and allocate extra computation to those queries that would benefit most from a more computationally intensive decoding procedure.
 
\section{Conclusion}
In this work, we propose a novel cost-efficient decoding method called Difficulty-Adaptive Self-Consistency (DSC), which fully leverages the difficulty information of given problem set to allocate computational resources based on their respective levels of difficulty, so as to alleviate issues of redundant sampling on simple questions and multiple re-sampling limitations that current cost-efficient self-consistency methods face. Extensive experiments show that DSC consistently surpasses two strong cost-effcient self-consistency baselines ASC and ESC by a significant margin on six popular benchmarks, while attaining  comparable accuracy. Additional experiments confirm the effectiveness of the proposed three-step algorithms.
%To obtain difficulty, we propose two strategies called candidates filtering and merge. Extensive experiments demonstrate that DSC notably boosts the performance on diverse range of tasks, exhibits superior robustness against noise in input responses, and can be generalized to those tasks where answer extraction is feasible through voting. Additional experiments confirm that the proposed candidates filtering and merge strategies can further enhance the performance of DSC while reducing the required computational cost.
\section*{Limitations}
Despite the remarkable efficiency gain on variety of reasoning tasks, the current implementation of DSC still suffers from the following two limitations:

\begin{itemize}

\item As the performance of difficulty ranking declines when handling mixed-type problems (Table ~\ref{tb:mix type}), DSC may encounter challenges when applied in scenarios requiring real-time reasoning for mixed-type problems. Exploring further classification of problems by type or optimizing the current Difficulty Ranking algorithm could enhance its applicability.
% DSC is difficult to apply directly in scenarios that require real-time reasoning for mixed-type problems. Further classification of problems by type or further optimization of the current Difficulty Ranking algorithm is needed.

\item Given that DSC demands an awareness of the test set to rank samples based on their difficulty, its use could be restricted in scenarios where a single user is only permitted one input at a time. Nevertheless, the application scenarios of DSC include but are not limited to: (1) The server end (such as OpenAI company, etc) simultaneously receives a large number of query requests from users. (2) A user possesses a batch of data that requires a one-time inference.
\end{itemize}
\section*{Ethics Statement}
All of the datasets used in this study were publicly available, and no annotators were employed for our data collection. We confirm that the datasets we used did not contain any harmful content and was consistent with their intended use (research). We have cited the datasets and relevant works used in this study.
\section*{Acknowledgements}
This work is supported by the Beijing Natural Science Foundation, China (Nos. 4222037, L181010).
\clearpage
% \section*{Acknowledgements}

\bibliography{anthology, custom}

\begin{thebibliography}{38}
\providecommand{\natexlab}[1]{#1}

\bibitem[{Agarwal et~al.(2024)Agarwal, Singh, Zhang, Bohnet, Rosias, Chan, Zhang, Faust, and Larochelle}]{agarwal2024many}
Rishabh Agarwal, Avi Singh, Lei~M Zhang, Bernd Bohnet, Luis Rosias, Stephanie~CY Chan, Biao Zhang, Aleksandra Faust, and Hugo Larochelle. 2024.
\newblock Many-shot in-context learning.
\newblock In \emph{ICML 2024 Workshop on In-Context Learning}.

\bibitem[{Aggarwal et~al.(2023)Aggarwal, Madaan, Yang et~al.}]{ASC}
Pranjal Aggarwal, Aman Madaan, Yiming Yang, et~al. 2023.
\newblock Let’s sample step by step: Adaptive-consistency for efficient reasoning and coding with llms.
\newblock In \emph{Proceedings of the 2023 Conference on Empirical Methods in Natural Language Processing}, pages 12375--12396.

\bibitem[{Brown et~al.(2024)Brown, Juravsky, Ehrlich, Clark, Le, R{\'e}, and Mirhoseini}]{brown2024large}
Bradley Brown, Jordan Juravsky, Ryan Ehrlich, Ronald Clark, Quoc~V Le, Christopher R{\'e}, and Azalia Mirhoseini. 2024.
\newblock Large language monkeys: Scaling inference compute with repeated sampling.
\newblock \emph{arXiv preprint arXiv:2407.21787}.

\bibitem[{Bubeck et~al.(2023)Bubeck, Chandrasekaran, Eldan, Gehrke, Horvitz, Kamar, Lee, Lee, Li, Lundberg et~al.}]{bubeck2023sparks}
S{\'e}bastien Bubeck, Varun Chandrasekaran, Ronen Eldan, Johannes Gehrke, Eric Horvitz, Ece Kamar, Peter Lee, Yin~Tat Lee, Yuanzhi Li, Scott Lundberg, et~al. 2023.
\newblock Sparks of artificial general intelligence: Early experiments with gpt-4.
\newblock \emph{arXiv preprint arXiv:2303.12712}.

\bibitem[{Chen et~al.(2024)Chen, Prasad, Saha, Stengel-Eskin, and Bansal}]{chen2024magicore}
Justin Chih-Yao Chen, Archiki Prasad, Swarnadeep Saha, Elias Stengel-Eskin, and Mohit Bansal. 2024.
\newblock Magicore: Multi-agent, iterative, coarse-to-fine refinement for reasoning.
\newblock \emph{arXiv preprint arXiv:2409.12147}.

\bibitem[{Cheng et~al.(2023)Cheng, Kasai, and Yu}]{cheng2023batch}
Zhoujun Cheng, Jungo Kasai, and Tao Yu. 2023.
\newblock Batch prompting: Efficient inference with large language model apis.
\newblock In \emph{Proceedings of the 2023 Conference on Empirical Methods in Natural Language Processing: Industry Track}, pages 792--810.

\bibitem[{Cobbe et~al.(2021)Cobbe, Kosaraju, Bavarian, Chen, Jun, Kaiser, Plappert, Tworek, Hilton, Nakano, Hesse, and Schulman}]{GSM8K}
Karl Cobbe, Vineet Kosaraju, Mohammad Bavarian, Mark Chen, Heewoo Jun, Lukasz Kaiser, Matthias Plappert, Jerry Tworek, Jacob Hilton, Reiichiro Nakano, Christopher Hesse, and John Schulman. 2021.
\newblock \href {https://arxiv.org/abs/2110.14168} {Training verifiers to solve math word problems}.
\newblock \emph{CoRR}, abs/2110.14168.

\bibitem[{Cong et~al.(2025)Cong, Qizhi, Zhao, and Yang}]{cong2025baton}
Peizhuang Cong, Chen Qizhi, Haochen Zhao, and Tong Yang. 2025.
\newblock Baton: Enhancing batch-wise inference efficiency for large language models via dynamic re-batching.
\newblock In \emph{THE WEB CONFERENCE 2025}.

\bibitem[{Damani et~al.(2024)Damani, Shenfeld, Peng, Bobu, and Andreas}]{damani2024learning}
Mehul Damani, Idan Shenfeld, Andi Peng, Andreea Bobu, and Jacob Andreas. 2024.
\newblock Learning how hard to think: Input-adaptive allocation of lm computation.
\newblock \emph{arXiv preprint arXiv:2410.04707}.

\bibitem[{Fu et~al.(2023)Fu, Peng, Ou, Sabharwal, and Khot}]{Specialize}
Yao Fu, Hao Peng, Litu Ou, Ashish Sabharwal, and Tushar Khot. 2023.
\newblock \href {https://doi.org/10.48550/arXiv.2301.12726} {Specializing smaller language models towards multi-step reasoning}.
\newblock \emph{CoRR}, abs/2301.12726.

\bibitem[{Geva et~al.(2021)Geva, Khashabi, Segal, Khot, Roth, and Berant}]{SQA}
Mor Geva, Daniel Khashabi, Elad Segal, Tushar Khot, Dan Roth, and Jonathan Berant. 2021.
\newblock \href {https://doi.org/10.1162/tacl\_a\_00370} {Did aristotle use a laptop? {A} question answering benchmark with implicit reasoning strategies}.
\newblock \emph{Trans. Assoc. Comput. Linguistics}, 9:346--361.

\bibitem[{Hendrycks et~al.(2021)Hendrycks, Burns, Kadavath, Arora, Basart, Tang, Song, and Steinhardt}]{MATH}
Dan Hendrycks, Collin Burns, Saurav Kadavath, Akul Arora, Steven Basart, Eric Tang, Dawn Song, and Jacob Steinhardt. 2021.
\newblock \href {https://datasets-benchmarks-proceedings.neurips.cc/paper/2021/hash/be83ab3ecd0db773eb2dc1b0a17836a1-Abstract-round2.html} {Measuring mathematical problem solving with the {MATH} dataset}.
\newblock In \emph{Proceedings of the Neural Information Processing Systems Track on Datasets and Benchmarks 1, NeurIPS Datasets and Benchmarks 2021, December 2021, virtual}.

\bibitem[{Hosseini et~al.(2014)Hosseini, Hajishirzi, Etzioni, and Kushman}]{AddSub}
Mohammad~Javad Hosseini, Hannaneh Hajishirzi, Oren Etzioni, and Nate Kushman. 2014.
\newblock \href {https://doi.org/10.3115/v1/d14-1058} {Learning to solve arithmetic word problems with verb categorization}.
\newblock In \emph{Proceedings of the 2014 Conference on Empirical Methods in Natural Language Processing, {EMNLP} 2014, October 25-29, 2014, Doha, Qatar, {A} meeting of SIGDAT, a Special Interest Group of the {ACL}}, pages 523--533. {ACL}.

\bibitem[{Jain et~al.(2023)Jain, Ma, Deoras, and Xiang}]{open}
Siddhartha Jain, Xiaofei Ma, Anoop Deoras, and Bing Xiang. 2023.
\newblock \href {https://doi.org/10.48550/arXiv.2307.06857} {Self-consistency for open-ended generations}.
\newblock \emph{CoRR}, abs/2307.06857.

\bibitem[{Jiang et~al.(2023)Jiang, Sablayrolles, Mensch, Bamford, Chaplot, Casas, Bressand, Lengyel, Lample, Saulnier et~al.}]{mistral}
Albert~Q Jiang, Alexandre Sablayrolles, Arthur Mensch, Chris Bamford, Devendra~Singh Chaplot, Diego de~las Casas, Florian Bressand, Gianna Lengyel, Guillaume Lample, Lucile Saulnier, et~al. 2023.
\newblock Mistral 7b.
\newblock \emph{arXiv preprint arXiv:2310.06825}.

\bibitem[{Kojima et~al.(2022)Kojima, Gu, Reid, Matsuo, and Iwasawa}]{ZeroCOT}
Takeshi Kojima, Shixiang~Shane Gu, Machel Reid, Yutaka Matsuo, and Yusuke Iwasawa. 2022.
\newblock \href {http://papers.nips.cc/paper\_files/paper/2022/hash/8bb0d291acd4acf06ef112099c16f326-Abstract-Conference.html} {Large language models are zero-shot reasoners}.
\newblock In \emph{NeurIPS}.

\bibitem[{Li et~al.(2024)Li, Yuan, Feng, Pan, Sun, Wang, Wang, and Li}]{li2024turning}
Yiwei Li, Peiwen Yuan, Shaoxiong Feng, Boyuan Pan, Bin Sun, Xinglin Wang, Heda Wang, and Kan Li. 2024.
\newblock Turning dust into gold: Distilling complex reasoning capabilities from llms by leveraging negative data.
\newblock In \emph{Proceedings of the AAAI Conference on Artificial Intelligence}, volume~38, pages 18591--18599.

\bibitem[{Li et~al.(2023)Li, Yuan, Feng, Pan, Wang, Sun, Wang, and Li}]{ESC}
Yiwei Li, Peiwen Yuan, Shaoxiong Feng, Boyuan Pan, Xinglin Wang, Bin Sun, Heda Wang, and Kan Li. 2023.
\newblock Escape sky-high cost: Early-stopping self-consistency for multi-step reasoning.
\newblock In \emph{The Twelfth International Conference on Learning Representations}.

\bibitem[{Lin et~al.(2024)Lin, Diesendruck, Du, and Abraham}]{linbatchprompt}
Jianzhe Lin, Maurice Diesendruck, Liang Du, and Robin Abraham. 2024.
\newblock Batchprompt: Accomplish more with less.
\newblock In \emph{The Twelfth International Conference on Learning Representations}.

\bibitem[{Liu et~al.(2024{\natexlab{a}})Liu, Yang, and Neville}]{liu2024cliqueparcel}
Jiayi Liu, Tinghan Yang, and Jennifer Neville. 2024{\natexlab{a}}.
\newblock Cliqueparcel: An approach for batching llm prompts that jointly optimizes efficiency and faithfulness.
\newblock \emph{arXiv preprint arXiv:2402.14833}.

\bibitem[{Liu et~al.(2024{\natexlab{b}})Liu, Lin, Hewitt, Paranjape, Bevilacqua, Petroni, and Liang}]{liu-etal-2024-lost}
Nelson~F. Liu, Kevin Lin, John Hewitt, Ashwin Paranjape, Michele Bevilacqua, Fabio Petroni, and Percy Liang. 2024{\natexlab{b}}.
\newblock \href {https://doi.org/10.1162/tacl_a_00638} {Lost in the middle: How language models use long contexts}.
\newblock \emph{Transactions of the Association for Computational Linguistics}, 11:157--173.

\bibitem[{Manvi et~al.(2024)Manvi, Singh, and Ermon}]{manvi2024adaptive}
Rohin Manvi, Anikait Singh, and Stefano Ermon. 2024.
\newblock Adaptive inference-time compute: Llms can predict if they can do better, even mid-generation.
\newblock \emph{arXiv preprint arXiv:2410.02725}.

\bibitem[{Miao et~al.(2020)Miao, Liang, and Su}]{asdiv}
Shen{-}Yun Miao, Chao{-}Chun Liang, and Keh{-}Yih Su. 2020.
\newblock \href {https://doi.org/10.18653/v1/2020.acl-main.92} {A diverse corpus for evaluating and developing english math word problem solvers}.
\newblock In \emph{Proceedings of the 58th Annual Meeting of the Association for Computational Linguistics, {ACL} 2020, Online, July 5-10, 2020}, pages 975--984. Association for Computational Linguistics.

\bibitem[{OpenAI(2023)}]{GPT4}
OpenAI. 2023.
\newblock \href {https://doi.org/10.48550/arXiv.2303.08774} {{GPT-4} technical report}.
\newblock \emph{CoRR}, abs/2303.08774.

\bibitem[{Patel et~al.(2021)Patel, Bhattamishra, and Goyal}]{svamp}
Arkil Patel, Satwik Bhattamishra, and Navin Goyal. 2021.
\newblock \href {https://doi.org/10.18653/v1/2021.naacl-main.168} {Are {NLP} models really able to solve simple math word problems?}
\newblock In \emph{Proceedings of the 2021 Conference of the North American Chapter of the Association for Computational Linguistics: Human Language Technologies, {NAACL-HLT} 2021, Online, June 6-11, 2021}, pages 2080--2094. Association for Computational Linguistics.

\bibitem[{Roy and Roth(2015)}]{multi}
Subhro Roy and Dan Roth. 2015.
\newblock \href {https://doi.org/10.18653/v1/d15-1202} {Solving general arithmetic word problems}.
\newblock In \emph{Proceedings of the 2015 Conference on Empirical Methods in Natural Language Processing, {EMNLP} 2015, Lisbon, Portugal, September 17-21, 2015}, pages 1743--1752. The Association for Computational Linguistics.

\bibitem[{Snell et~al.(2024)Snell, Lee, Xu, and Kumar}]{snell2024scaling}
Charlie Snell, Jaehoon Lee, Kelvin Xu, and Aviral Kumar. 2024.
\newblock Scaling llm test-time compute optimally can be more effective than scaling model parameters.
\newblock \emph{arXiv preprint arXiv:2408.03314}.

\bibitem[{Son et~al.(2024)Son, Baek, Nam, Jeong, and Kim}]{son-etal-2024-multi-task}
Guijin Son, SangWon Baek, Sangdae Nam, Ilgyun Jeong, and Seungone Kim. 2024.
\newblock \href {https://doi.org/10.18653/v1/2024.acl-long.304} {Multi-task inference: Can large language models follow multiple instructions at once?}
\newblock In \emph{Proceedings of the 62nd Annual Meeting of the Association for Computational Linguistics (Volume 1: Long Papers)}, pages 5606--5627, Bangkok, Thailand. Association for Computational Linguistics.

\bibitem[{Talmor et~al.(2019)Talmor, Herzig, Lourie, and Berant}]{CSQA}
Alon Talmor, Jonathan Herzig, Nicholas Lourie, and Jonathan Berant. 2019.
\newblock \href {https://doi.org/10.18653/v1/n19-1421} {Commonsenseqa: {A} question answering challenge targeting commonsense knowledge}.
\newblock In \emph{Proceedings of the 2019 Conference of the North American Chapter of the Association for Computational Linguistics: Human Language Technologies, {NAACL-HLT} 2019, Minneapolis, MN, USA, June 2-7, 2019, Volume 1 (Long and Short Papers)}, pages 4149--4158. Association for Computational Linguistics.

\bibitem[{Wang et~al.(2024)Wang, Li, Feng, Yuan, Pan, Wang, Hu, and Li}]{FSC}
Xinglin Wang, Yiwei Li, Shaoxiong Feng, Peiwen Yuan, Boyuan Pan, Heda Wang, Yao Hu, and Kan Li. 2024.
\newblock Integrate the essence and eliminate the dross: Fine-grained self-consistency for free-form language generation.
\newblock In \emph{Proceedings of the 62nd Annual Meeting of the Association for Computational Linguistics (Volume 1: Long Papers)}, pages 11782--11794.

\bibitem[{Wang et~al.(2022)Wang, Wei, Schuurmans, Le, Chi, Narang, Chowdhery, and Zhou}]{wang2022self}
Xuezhi Wang, Jason Wei, Dale Schuurmans, Quoc~V Le, Ed~H Chi, Sharan Narang, Aakanksha Chowdhery, and Denny Zhou. 2022.
\newblock Self-consistency improves chain of thought reasoning in language models.
\newblock In \emph{The Eleventh International Conference on Learning Representations}.

\bibitem[{Wang et~al.(2023)Wang, Wei, Schuurmans, Le, Chi, Narang, Chowdhery, and Zhou}]{SC}
Xuezhi Wang, Jason Wei, Dale Schuurmans, Quoc~V. Le, Ed~H. Chi, Sharan Narang, Aakanksha Chowdhery, and Denny Zhou. 2023.
\newblock \href {https://openreview.net/pdf?id=1PL1NIMMrw} {Self-consistency improves chain of thought reasoning in language models}.
\newblock In \emph{The Eleventh International Conference on Learning Representations, {ICLR} 2023, Kigali, Rwanda, May 1-5, 2023}. OpenReview.net.

\bibitem[{Wei et~al.(2022{\natexlab{a}})Wei, Wang, Schuurmans, Bosma, Ichter, Xia, Chi, Le, and Zhou}]{COT}
Jason Wei, Xuezhi Wang, Dale Schuurmans, Maarten Bosma, Brian Ichter, Fei Xia, Ed~H. Chi, Quoc~V. Le, and Denny Zhou. 2022{\natexlab{a}}.
\newblock \href {http://papers.nips.cc/paper\_files/paper/2022/hash/9d5609613524ecf4f15af0f7b31abca4-Abstract-Conference.html} {Chain-of-thought prompting elicits reasoning in large language models}.
\newblock In \emph{NeurIPS}.

\bibitem[{Wei et~al.(2022{\natexlab{b}})Wei, Wang, Schuurmans, Bosma, Xia, Chi, Le, Zhou et~al.}]{wei2022chain}
Jason Wei, Xuezhi Wang, Dale Schuurmans, Maarten Bosma, Fei Xia, Ed~Chi, Quoc~V Le, Denny Zhou, et~al. 2022{\natexlab{b}}.
\newblock Chain-of-thought prompting elicits reasoning in large language models.
\newblock \emph{Advances in neural information processing systems}, 35:24824--24837.

\bibitem[{Wu et~al.(2024{\natexlab{a}})Wu, Sun, Li, Welleck, and Yang}]{wu2024empirical}
Yangzhen Wu, Zhiqing Sun, Shanda Li, Sean Welleck, and Yiming Yang. 2024{\natexlab{a}}.
\newblock An empirical analysis of compute-optimal inference for problem-solving with language models.
\newblock \emph{URL https://arxiv. org/abs/2408.00724}.

\bibitem[{Wu et~al.(2024{\natexlab{b}})Wu, Zeng, Zhang, Tan, Shen, and Jiang}]{wu2024large}
Zhenyu Wu, Qingkai Zeng, Zhihan Zhang, Zhaoxuan Tan, Chao Shen, and Meng Jiang. 2024{\natexlab{b}}.
\newblock Large language models can self-correct with key condition verification.
\newblock In \emph{Proceedings of the 2024 Conference on Empirical Methods in Natural Language Processing}, pages 12846--12867.

\bibitem[{Zheng et~al.(2023)Zheng, Liu, Xie, Li, and Li}]{PHP}
Chuanyang Zheng, Zhengying Liu, Enze Xie, Zhenguo Li, and Yu~Li. 2023.
\newblock \href {https://doi.org/10.48550/arXiv.2304.09797} {Progressive-hint prompting improves reasoning in large language models}.
\newblock \emph{CoRR}, abs/2304.09797.

\bibitem[{Zhou et~al.(2023)Zhou, Sch{\"{a}}rli, Hou, Wei, Scales, Wang, Schuurmans, Cui, Bousquet, Le, and Chi}]{LeastToMost}
Denny Zhou, Nathanael Sch{\"{a}}rli, Le~Hou, Jason Wei, Nathan Scales, Xuezhi Wang, Dale Schuurmans, Claire Cui, Olivier Bousquet, Quoc~V. Le, and Ed~H. Chi. 2023.
\newblock \href {https://openreview.net/pdf?id=WZH7099tgfM} {Least-to-most prompting enables complex reasoning in large language models}.
\newblock In \emph{The Eleventh International Conference on Learning Representations, {ICLR} 2023, Kigali, Rwanda, May 1-5, 2023}. OpenReview.net.

\end{thebibliography}
% \bibliography{custom}

\clearpage

\appendix
\section{Appendix}

\subsection{Prompt for Difficulty Ranking}
\label{prompt}
\begin{quote}
{\itshape
Your task is to rank the given questions from easy to hard based on their difficulty level. Questions to be evaluated:

Q1:\{Question 1\}

Q2:\{Question 2\}

...

Qn: \{Question n\}

The output format should be a comma-separated list containing the Q{number of corresponding question}. Do not give any explanation. 

Difficulty Ranking result (from easy to hard):
}
\end{quote}

% citep{mistral}

\subsection{Comparison of Inference Time Between Different Methods}
Considering the inference time is important for real world scenarios, we calculate the inference time of different methods on the MATH test set (5000 questions) with GPT-4. As shown in Table \ref{tb:inference time}, for DSC, the ranking time corresponds to the time produced by step 1, while sampling time corresponds to the time generated by step 2 and step 3. The results show that, compared to SC, both ASC and ESC require significantly more time due to the need for a large amount of resampling. DSC, on the other hand, effectively mitigates this issue by pre-allocating the sample size for questions.

\subsection{Comparison with ESC under Different Window Size}
To further demonstrate the effectiveness of DSC compared to ESC, we make a comparison between the performance of DSC and ESC under different window sizes on the MATH dataset using GPT-4, while maintaining the rest of the settings completely consistent with the main experiment. As shown in Table \ref{tb:ESC with various window size}, when the window size is greater than or equal to 5, ESC can maintain its Accuracy and its cost increases as the window size enlarges. However, when the window size is less than 5, the Accuracy of ESC drops significantly due to excessively relaxed constraints. Overall, the performance of DSC consistently surpasses that of ESC by a large margin.

\subsection{Performance of Difficulty Ranking under Mixing Types of Questions}
To explore whether Difficulty Ranking can effectively rank problems of different types in terms of difficulty, we compare the performance of Difficulty Ranking under mixed and unmixed problem types on MATH, as shown in Table \ref{tb:mix type}. Specifically, for unmixed, we perform Difficulty Ranking on seven subsets separately; for mixed, we directly execute Difficulty Ranking on random shuffled MATH dataset, and calculate the correlation on the entire dataset for both. The results indicate that mixing different types of questions for LLM to rank can lead to a decrease in performance. This might be due to the challenge for LLM in adhering to the same assessment standards for various types of questions.\footnote{It could also lead to inaccurate ranking for humans.}

\begin{table}[t]
\setlength{\tabcolsep}{3.30pt}
\small
\centering
\begin{tabular}{lccc}
\toprule
Time (hours) & Ranking Time & Sampling Time & Total Time \\ \toprule
SC & 0 & 11.75 & 11.75 \\
ASC & 0 & 221 & 221 \\
ESC & 0 & 57.26 & 57.26 \\
DSC & 2.26 & 17.08 & 19.34 \\ \bottomrule
\end{tabular}
\caption{Inference time analysis of different methods on MATH using GPT-4.}
\label{tb:inference time}
\end{table}

\begin{table}[t]
\setlength{\tabcolsep}{2.90pt}
\small
\centering
\begin{tabular}{lcccccc}
\toprule
Window & 3 & 4 & 5 & 6 & 7  & DSC \\ \toprule
Acc & 58.10 & 58.40 & 58.49 & 58.51 & 58.51 & 58.51  \\
Cost & 0.3505 & 0.3905 & 0.4062 & 0.4135 & 0.4173 & 0.3142 \\ \bottomrule
\end{tabular}
\caption{Comparison of ESC under different window size and DSC on MATH with GPT-4.}
\label{tb:ESC with various window size}
\end{table}

\begin{table}[t]
\renewcommand{\arraystretch}{0.94}
\centering
\small
\begin{tabular}{lccc}
\toprule
Model & Metric & Unmixed & Mixed \\ \toprule
\multirow{3}{*}{GPT-4} & Spearman & 0.6451 & 0.5148 \\
 & Pearson & 0.6487 & 0.5197 \\
 & Kendall & 0.5060 & 0.3931 \\ \midrule
\multirow{3}{*}{GPT-3.5-Turbo} & Spearman & 0.4828 & 0.4070 \\
 & Pearson & 0.4887 & 0.4145 \\
 & Kendall & 0.3697 & 0.3073 \\ \bottomrule
\end{tabular}
\caption{Performance comparison of Difficulty Ranking under mixed and unmixed problem types on MATH.}
\label{tb:mix type}
\end{table}

% \usepackage{multirow}
\begin{table}[t]
\renewcommand{\arraystretch}{0.94}
\centering
\small
\begin{tabular}{lccccc}
\toprule
\multirow{2}{*}{Setting} & \multicolumn{5}{c}{N} \\ \cmidrule{2-6} 
 & 1 & 5 & 10 & 20 & 40 \\ \midrule
Zero-shot & 27.01 & 31.44 & 33.84 & 35.21 & 35.93 \\
Few-shot & 33.08 & 38.54 & 41.39 & 43.01 & 43.94 \\ \bottomrule
\end{tabular}
\caption{Accuracy Comparison of SC under Zero-Shot and Few-Shot Settings with Different Sample Sizes N on MATH500 Using GPT-3.5-Turbo}
\label{tb:few-shot}
\end{table}

\subsection{Hyperparameter Experiment of Difficulty Ranking}
\begin{figure*}[t]
\begin{center}
\includegraphics[width=0.90\textwidth]{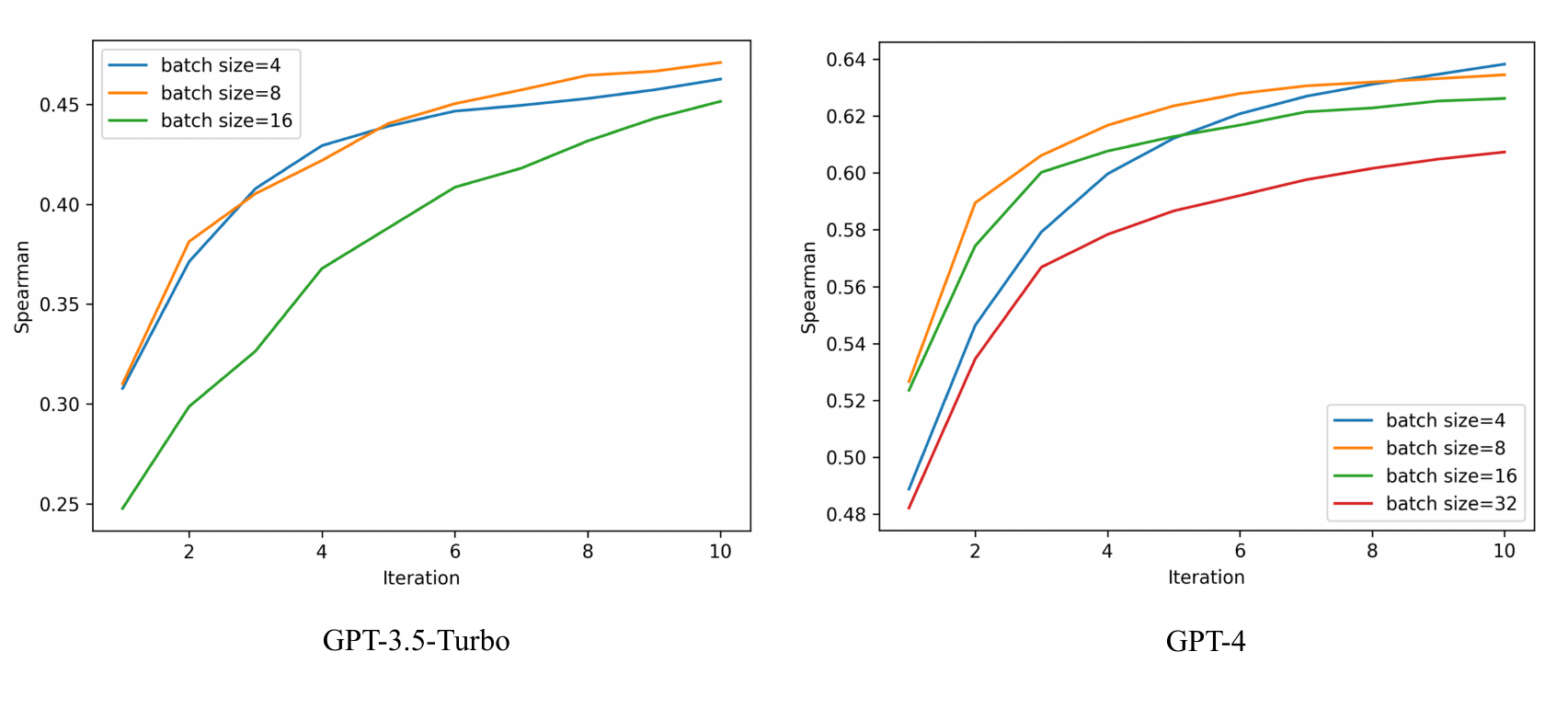}
\end{center}
\caption{Performance of Difficulty Ranking under different batch sizes and iterations on MATH with GPT-3.5-Turbo and GPT-4.}
\label{fig:hyper}
\end{figure*}

As shown in Figure \ref{fig:hyper}, we conduct hyperparameter experiments on the proposed Difficulty Ranking algorithm. The experimental results indicate that a large batch size ($ \geq 16 $) leads to a decrease in LLM ranking performance, which is consistent with our expectations. For both GPT-3.5-Turbo and GPT-4, difficulty ranking approximately converges in the 5th round. Therefore, we choose 5 and 8 as the default values for iteration and batch size for Difficulty Ranking, respectively.

\subsection{Necessity of Few-Shot Setting in DSC}

Previous studies \citep{COT, SC} have experimentally demonstrated that compared to zero-shot (without demonstrations), few-shot (with demonstrations) significantly improves the performance of LLMs on reasoning tasks. Therefore, we adopted the same few-shot setting as SC, ASC, and ESC. To further verify this, we test the Accuracy of SC on GPT-3.5-Turbo under both zero-shot and few-shot (8-shot as default) settings with different sample sizes N on MATH500. 

The results (Table \ref{tb:few-shot}) show that removing demonstrations leads to a significant performance drop, validating the necessity of multiple demonstrations. As a matter of fact, during our research, we tried to dynamically allocate the number of demonstrations based on question difficulty to reduce input costs. However, we find that reducing the number of demonstrations significantly decreased performance, regardless of the difficulty of the questions.

Moreover, recent study \citep{agarwal2024many} has demonstrated that a greater number of demonstrations generally leads to more performance improvements. For most tasks, the optimal number of demonstrations is often greater than 512, far exceeding the default setting of 8 in DSC, which results in much higher input costs. This further highlights the value of DSC.

\subsection{Comparison of DSC with Self-Correction and Verifier-Based Methods}

We consider that self-correct \citep{wu2024large}, verifier-based methods \citep{wu2024empirical}, and Self-consistency \citep{SC} can all be seen as approaches that aim to improve model performance by increasing computational effort, albeit through different implementation paths. In comparison, DSC takes a different direction by focusing on optimizing efficiency—reducing computational costs as much as possible while maintaining model performance. This distinction positions DSC as complementary to self-correct, verifier-based methods, and Self-consistency, with a unique focus on efficiency.

Furthermore, while DSC is implemented based on Self-consistency, its concept of dynamically allocating computational resources according to question difficulty could also be adapted to self-correct and verifier-based methods to enhance efficiency. For example, in the case of self-correct, this might involve assessing whether a query requires self-correction based on its difficulty and estimating the necessary degree of correction (e.g., the number of iterations). Similarly, for verifier-based methods, which already involve multiple sampling followed by verification, integrating DSC could potentially lead to more efficient inference processes.

\subsection{Background of ESC and ASC}
\label{sec: background}

% Considering that our approach involves the use of ESC and ASC related concepts, we provide a formal introduction to their methods here for better understanding.

\paragraph{ESC}
ESC proposes extension window sampling, where the extension window is a fixed preset value $w$, Each time, $w$ entries are sampled through the LLM and added to the sampling set. If the answers of the current 
$w$ samples are the same or the preset maximum sampling value is reached, the sampling is stopped. The core idea of ESC is that if the model's one-time $w$ sample answers are completely the same, it can be considered that it has a high confidence in this answer, and sampling can be stopped.

\paragraph{ASC}
ASC \citep{ASC} proposes the Dirichlet Stopping Criteria, where sampling is conducted one by one. After each sampling, the Dirichlet Stopping Criteria is used to determine whether all current samples meet a specific distribution. If they do, the sampling is stopped. If not, the one-by-one sampling continues until the preset maximum sampling value is reached. The Dirichlet Stopping Criteria is shown in Equation \ref{eq:asc}, where $v$ represents the current set of samples, $m$ is the number of $v$, $C_{thresh}$ is the preset threshold (set to 0.95 for ASC in default), and $p_1$ is the probability of the most frequently occurring answer in the set $v$. The core idea of ASC is to measure the two answers with the highest probability in the current set of samples. If the difference between the probability of the answer with the highest probability and the second highest probability exceeds a threshold ($C_{thresh}$), it can be considered that the model is very confident in the answer with the highest probability, and sampling can be stopped. For further details, please refer to the original paper.

\begin{equation}
P\left(p_1>\max _{i=2}^m p_i \mid v\right)>C_{\text {thresh }}
\label{eq:asc}
\end{equation}

\subsection{Application Scenarios of DSC}
As discussed in the Limitations section, DSC could be constrained in scenarios where a single user is restricted to one input at a time. However, it is widely applicable to common scenarios including: (1) servers (e.g., OpenAI or similar providers) managing a large number of simultaneous query requests from different users (e.g., ChatGPT receives requests from multiple users to solve multiple math problems simultaneously), and (2) cases where a user perform inference on a batch of data (E.g., Submitting multiple requests at once to efficiently receive feedback).

For these application scenarios, Recently, an increasing number of studies have focused on performing inference on large volumes of samples with LLMs \citep{cheng2023batch, linbatchprompt, son-etal-2024-multi-task, liu2024cliqueparcel, cong2025baton}, which is computationally and financially costly in industrial and real-world applications. Therefore, the application scenarios of DSC have become a key focus in current research, attracting extensive attention from the academic and industrial communities, and we believe DSC's potential applications and value will be more and more substantial in the future.

% To further illustrate the wide range of real-world applications for DSC, we validate its effectiveness when handling small batches of queries (Considering that typical users usually deal with a limited number of queries rather than the scale of an entire dataset). We randomly divided all the datasets into multiple subsets of a specified batch size instead of processing all queries simultaneously. Following \citet{linbatchprompt}, we set the batch size to 32, allowing DSC to only access the current batch of 32 queries instead of the entire dataset. The judge window size was set to 8, and the prediction window size to 5, while all other experimental settings remained consistent with the main experiments. The experimental results are as follows:

% Our results show that DSC remains effective under the constraint of small-batch queries, indicating that DSC can be applied to a broader range of real-world scenarios.

\end{document}